\documentclass[10pt,twocolumn,letterpaper]{article}

\usepackage{wacv}
\usepackage{times}
\usepackage{epsfig}
\usepackage{graphicx}
\usepackage{amsmath}
\usepackage{amssymb}
\usepackage{float}

\wacvfinalcopy 

\def\wacvPaperID{616} 

\ifwacvfinal\fi
\begin{document}

\title{Size to Depth: A New Perspective for Single Image Estimation}

\author{Yiran Wu \hspace{2cm} Sihao Ying \hspace{2cm} Lianmin Zheng\\
Shanghai Jiao Tong University\\
{\tt\small \{yiranwu,}
{\tt\small yingsihao,}
{\tt\small mercy\_zheng\}@sjtu.edu.cn}
}

\maketitle
\ifwacvfinal\thispagestyle{empty}\fi

\begin{abstract}
   
In this paper we consider the problem of single monocular image depth estimation. It is a challenging problem due to its ill-posedness nature and has found wide application in industry. Previous efforts belongs roughly to two families: learning-based method and interactive method. Learning-based method, in which deep convolutional neural network (CNN) is widely used, can achieve good result. But they suffer low generalization ability and typically perform poorly for unfamiliar scenes. Besides, data-hungry nature for such method makes data aquisition expensive and time-consuming. Interactive method requires human annotation of depth which, however, is errorneous and of large variance.

To overcome these problems, we propose a new perspective for single monocular image depth estimation problem: size to depth. Our method require sparse label for real-world size of object rather than raw depth. A Coarse depth map is then inferred following geometric relationships according to size labels. Then we refine the depth map by doing energy function optimization on
conditional random field(CRF). We experimentally demonstrate that our method outperforms traditional depth-labeling methods and can produce satisfactory depth maps.
\end{abstract}

\section{Introduction}

Single image depth estimation is a fundamental problem which has found wide applications in computer vision. It is a challenging problem for its ill-posedness due to the inherent ambiguity of single monocular image. As shown in Figure 1, according to the property of projection, object in the image can locate at both the near plane or the far plane. Every position along the projection direction is a possible location of the according object. While each possible position produces identical image, their distance to the camera (the depth) vary over a wide range. The above example demonstrates the difficulty of single image depth estimation task. Following the same reason, there are infinitely many possible scenes that correspond to certain image, each with different depth maps. Such phenomena accounts for the strong ill-posedness of the depth estimation problem.

\begin{figure}[ht]
\begin{center}
\fbox{\rule{0pt}{0in} \rule{0\linewidth}{0pt}
   \includegraphics[width=0.8\linewidth]{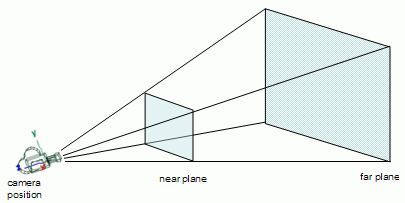}}
\end{center}
   \caption{Ambiguity of depth in single image.}
\label{fig:long}
\label{fig:onecol}
\end{figure}

For images of real scene, object prior is a clue to resolve the ambiguity of depth. While the object plane has infinite possible position, many of them correspond to impractical real-world size of objects. The basic equation in computer vision
\begin{equation}
\text{depth} \propto \text{focal\ length} * \frac{\text{real-world\ size}} {\text{size\ in\ photo}}
\end{equation}
implies that for fixed focal length and object size in photo (which is the case for single image depth estimation), depth of object is proportional to its real-world size. Through common knowledge and daily experience, we can extract some range so that most time size of the object is within such range. From the equation above, bounding the size of the object also gives a (typically tight) bound to its depth and thus provides strong information for depth. It implies that object size is of fundamental importance in depth estimation.

Some previous works try to inject object prior by having human label depth directly. This kind of method typically labels several carefully-selected characteristic area. And depth of the rest region is worked out by propagation from labeled area. Such method fail to give satisfying results. Their failure lies mainly on the insensitivity of human to the absolute value of depth. Human are not good at labeling raw depth. To demonstrate our claim, we perform the simple experiment below: take 5 images with ground truth depth (collected by Kinect), randomly choose 10 pixels in every image and let 10 people label the raw depth of each pixel, compare the result with the ground truth. The final result shows roughly 21\% relative error which is far from accurate. Besides, long hesitation is observed before deciding the final depth, which indicates that it is also inefficient. So we make the conclusion that labeling raw depth is inadvisable.

Inspired by equation (1), we propose a method to label size instead. According to equation (1), for single image where focal length and object size in photo is fixed, depth is proportional to size and are thus equivalent. If we can get size by following the labeling-propagation process, relative depth is also known and absolute depth can be readily calculated given camera parameters. We find that labeling size can be much more accurate and efficient. Experimental result supports us: taking picture of common objects such as shelves, desks, chairs, measure their size, and let 10 people label the size, relative error decreases all the way down to 8\%, and most testers give their answer instantly. Thus conclusion can be made that labeling size is a better method for depth estimation, with higher efficiency and accuracy. 

Despite the advantage of labeling depth, several challenges still remain to be tackled. The first problem is to decide the target to label. When labeling raw depth, pixel is a reasonable target to be labeled. But in our size-to-depth formulation, clue of depth comes from objects which is a semantic unit. It does not make sense to divide them into pixels. In addition, picking a representative pixel for an object is not trivial. But treating object as a whole necessitates the definition of object boundary. If we ask human to draw the boundary, it will be too time-consuming and the boundary will be noisy. We design a patch-to-size method to handle the problem with flexibility.

Propagation method is another concern. It has to figure out a way to preserve depth difference on object boundaries while remain the local connectivity of the depth map. The propagation method also need to tradeoff between fine-grained but computationally extensive high-resolution and coarse resolution with less computation expense. We apply conditional random field (CRF) to do the propagation.

To sum up, we highlight the main contributions of this work as follows:
\begin{itemize}
\item We propose a new perspective for single monocular image depth estimation problem: from size to depth. Given manually labeled size for some objects in image, we are able to infer relative depth through simple geometric relationships. 
\item We design a special conditional random field (CRF) model to apply depth propagation and finally generate the whole depth map.
\item We demonstrate the superiorities of our method compared to the traditional depth-labeling method.
\end{itemize}


\section{Related works}

Increasing number of methods are trying to estimate depth for single monocular image, which can be roughly classified into two families: learning-based method and interactive method.

Some traditional learning-based methods formulate depth estimation as a markov random field (MRF) learning problem. Saxena \etal \cite{NIPS2005_2921} use linear regression and MRF to predict depth from a set of image feature. Liu \etal \cite{Liu_2014_CVPR} propose a discrete-continuous conditional random field (CRF) model to take relations between adjacent superpixels into consideration. Realizing the strong correlation between depth estimation and semantic segmentation, Liu \etal \cite{Liu+al:CVPR10} make use of predicted semantic labels to guide 3D reconstruction by enforcing depth related to class and geometry prior.

Most of recent learning-based approaches rely on the application of deep learning, among which deep convolutional neural network (CNN) is used most commonly. Eigen \etal \cite{DBLP:journals/corr/EigenPF14} design a global coarse-scale deep CNN to regress a rough depth map directly from an input image. They then train a local fine-scale network to make local refinements. Liu \etal \cite{Liu_2015_CVPR} propose a deep convolutional neural field model for depth estimation by exploring CNN and  CRF. They jointly learn the unary and pairwise potentials of CRF in a unified deep CNN framework. Like in traditional learning-based method, Wang \etal \cite{Wang_2015_CVPR} use a deep CNN to jointly predict a global layout composed of pixel-wise depth values as well as semantic labels, and improved performance by allowing interactions between depth and semantic information. 

Compared to our method, learning-based methods require plenty of ground truth data and are not able to generalize to unseen images outside the dataset. Some of the methods \cite{Liu+al:CVPR10, Wang_2015_CVPR} need result of segmentation algorithm.

Human interactive methods exploit human ability to interpret 2D images, using human annotation. Some previous works make use of geometric elements (lines or plains) in images to predict depth. Criminisi \etal \cite{Criminisi2000} describe a way to compute 3D affine measurements given human inputs providing geometric information determined from the image. Later works \cite{Lee2009GeometricRF} by Lee \etal try to generate plausible interpretations of a scene from a collection of line segments automatically extracted from a single indoor image. These methods are limited for requiring a large amount of straight lines or plains in the image to provide enough evidences for 3D structure inference. Lopez \etal \cite{ceig.20141109} formulate the problem as an optimization process by assuming that image regions with low gradients will have similar depth values. In their method, depth values are propagated between pixels with small image gradients under a number of human-defined constraints. Our method, In contrast, requires only size information, which is more trivial for human to label than geometric or depth information. We will show this in the following sections.

\section{Our method}
In this section, we present details of our size-to-depth algorithm for depth estimation. We use bold upper case letter to denote matrices and bold lower case letter to denote vectors.

\subsection{Overview}
Our goal is to estimate pixelwise depth for single image of general cases. As directly labeling depth in numeric value is inefficient and inaccurate, we propose the quick and precise size-to-depth algorithm.

Following the labeling-propagation pipeline, the first step of our algorithm is labeling size which is an fast and efficient technique. According to equation (1), real-world size is proportional to depth in single image depth estimation setting. Thus size is equivalent to relative depth and we can calculate absolute depth given the camera parameters.

After patch-to-size step, we obtain size for each patch (which is equivalent to depth). But the result is rather coarse with rigid boundary. Another depth refinement step is introduced to smooth depth gaps and interpolate between pixels under the constraint of depth annotation.

The depth refinement step is formulated as an energy function optimization problem in conditional random field (CRF). Mathematically, let $\mathbf{x}$ be the RGB image, $\mathbf{y}$ be the corresponding depth, we model the conditional probability distribution of data with
\begin{equation}
\text{Pr}(\mathbf{y}|\mathbf{x}) = \frac{1}{Z(\mathbf{x})}\text{exp}(-E(\mathbf{x}, \mathbf{y})),
\end{equation}
where $E(\mathbf{x}, \mathbf{y})$ is the energy function and $Z(\mathbf{x})$ is a normalization term given by
\begin{equation}
Z(\mathbf{x}) = \int_{\mathbf{y}}\text{exp}(-E(\mathbf{x}, \mathbf{y})) d\mathbf{y}.
\end{equation}
Maximum a posteriori (MAP) solution $\mathbf{y}^*$ gives the depth of maximum probability $\mathbf{y}$ for observed image $\mathbf{x}$ which is the best estimation of depth
\begin{equation}
\mathbf{y^*} = \text{argmax}_{\mathbf{y}} \log \text{Pr}(\mathbf{y}|\mathbf{x}).
\end{equation}
\par
In the following, we will give a detailed discussion about components of our algorithm.
\subsection{Annotation formulation}
Realizing the difficulty to model object knowledge and spatial relation, we introduce human annotation for additional information. Due to the urge to preserve the structure of sematic unit and the difficulty to pick representative pixel for certain object, we discard the scheme to label size at pixel level. But getting accurate object boundary requires extensive annotation labor or the inclusion of segmentation which will greatly degrade convenience and time complexity. We propose a patch-to-size method that tradeoff between the two cases and partially preserves structure of semantic unit while remains fast and convenient. In this formulation, image is divided into grids of equally sized patches. And human are asked to label real-world size of dominant component in each patch. The dominant component treatment keeps the structure of semantic units. It enables the semantically homogeneous pixels to be evenly constrained, which reduces computational cost. Meanwhile, dominant component does not deterministically specify certain object, avoiding the inconvenience and inefficiency of annotation introduced by the problem of defining object boundary. However, the obvious disadvantage of introducing dominant component is one have to wisely select the proper dominant component which requires certain level of experience.

We make the general assumption that depth of dominant component in the patch is representative of the entire patch (this does not mean we will assign same depth for every pixel in the patch, actually we are assigning same constraint, discussed later in Section 3.4). The assumption sounds inapplicable in some cases such as images depicting a person in the background of sky where depth in same patch may vary greatly. But it can be fixed by introducing CRF as shown in section 3.4.

\begin{figure*}[!ht]
\begin{center}
\fbox{\rule{0pt}{0in} \rule{0\linewidth}{0pt}
\includegraphics[width=17.5cm]{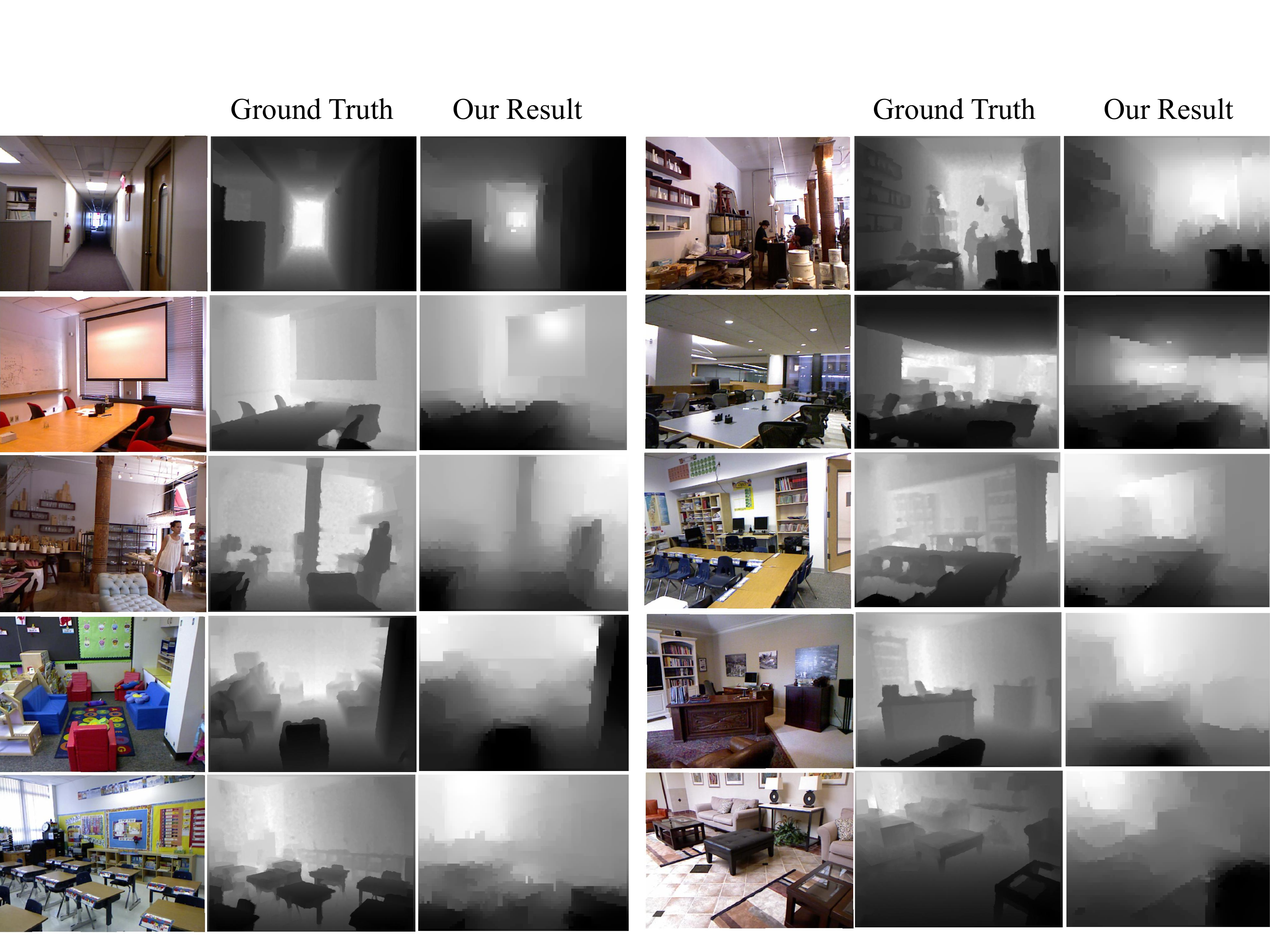}}
\end{center}
   \caption{Results of our method on NYU Depth dataset. The left column contains original images. Ground truth depth maps are placed in the center column and our results are presented in the right column.}
\label{fig:short}
\end{figure*} 

\subsection{Conditional random field}
As is typically formulated, energy function of our CRF consists of unary and binary terms.
\begin{align}
E(\mathbf{x}, \mathbf{y}) &= \sum_{p \in P}E_{unary}(\mathbf{y}_{p}, \mathbf{x}) \notag \\ &\quad + \lambda \sum_{(b, c) \in A} \frac{1}{2} E_{binary}(\mathbf{y}_b, \mathbf{y}_c, \mathbf{x}),
\end{align}
where $P$ is the set of pixels in the image, $A$ is set of 4-connected adjacent pair of pixels, $E_{unary}\text{ and }E_{binary}$ are relatively unary and binary terms, and $\lambda$ is a hyperparameter controlling the tradeoff of unary and binary term.

The unary term is given by
\begin{align}
E_{unary}(\mathbf{y}_{p}, \mathbf{x}) &= (\mathbf{y}_p - \mathbf{d}_p)^2,
\end{align}
where $\mathbf{d}$ is the depth annotation. This term basically forces the depth of image to match with labeled depth of its belonging patches. It will constrain the global layout of depth map to be relatively reasonable. 

\par
The binary term is given by
\begin{equation}
E_{binary}(\mathbf{y}_b, \mathbf{y}_c, \mathbf{x}) = \text{sim}(b, c) (\mathbf{y}_b-\mathbf{y}_c)^2,
\end{equation}
where $\text{sim}(b, c)$ is a similarity function of pixel. This term penalizes gap of depth between neighboring pixels which encourages local continuity of the depth field.
Similarity function between pixels can be represented by image gradient
\begin{equation}
\text{sim}(b, c) = e^{-\beta |I_b-I_c|},
\end{equation}
where $I_b, I_c$ denotes image intensity at b, c, and hyper-parameter $\beta$ controls the strength of local continuity. Smaller $\beta$ corresponds to stronger continuity constraint.
\subsection{MAP in CRF}
In the case of manual annotation, depth $\mathbf{d}$ is predefined by user. And if we measure similarity by image gradient, energy function can be written as
\begin{align}
E(\mathbf{x}, \mathbf{y}) &= \sum_{p \in P} (\mathbf{y}_p-\mathbf{d}_p)^2 + \lambda \sum_{b, c \in A}\text{sim}(b, c)(\mathbf{y}_b-\mathbf{y}_c)^2 \notag \\
&= \mathbf{y}^T (\ I+\lambda(\ \text{diag}(\mathbf{s}) - W )\ )\mathbf{y}-2\mathbf{y}^T\mathbf{d}+\mathbf{d}^T\mathbf{d}  \notag
\\ &= \mathbf{y}^T W'\mathbf{y}-2\mathbf{y}^T\mathbf{d}+\mathbf{d}^T\mathbf{d} 
\end{align}
where $I$ is the identity matrix, $\mathbf{d}$ is the annotation vector, $W_{ij} = \text{sim}(i, j) $, $\mathbf{s}_i = \sum_j W_{ij}$ and
$W' = I+\lambda(\ \text{diag}(\mathbf{s}) - W ).$
\begin{equation}
\text{Pr}(\mathbf{y}|\mathbf{x}) = \frac{\mathbf{y}^T W'\mathbf{y}-2\mathbf{y}^T\mathbf{d}+\mathbf{d}^T\mathbf{d} }{Z(\mathbf{x})}
\end{equation}
The optimization is performed by taking gradient w.r.t $\mathbf{y}$.
\begin{align}
\frac{\partial{\text{Pr}(\mathbf{y}|\mathbf{x})}}{\partial\mathbf{y}} &= \frac{\mathbf{y} ^T(W'+W'^T)-2\mathbf{d}^T}{Z(\mathbf{x})} = 0 \notag \\
\mathbf{y} &= 2(W'+W'^T)^{-1}\mathbf{d}.
\end{align}

\begin{figure*}[!ht]
\begin{center}
\fbox{\rule{0pt}{0in} \rule{0\linewidth}{0pt}
\includegraphics[width=15cm]{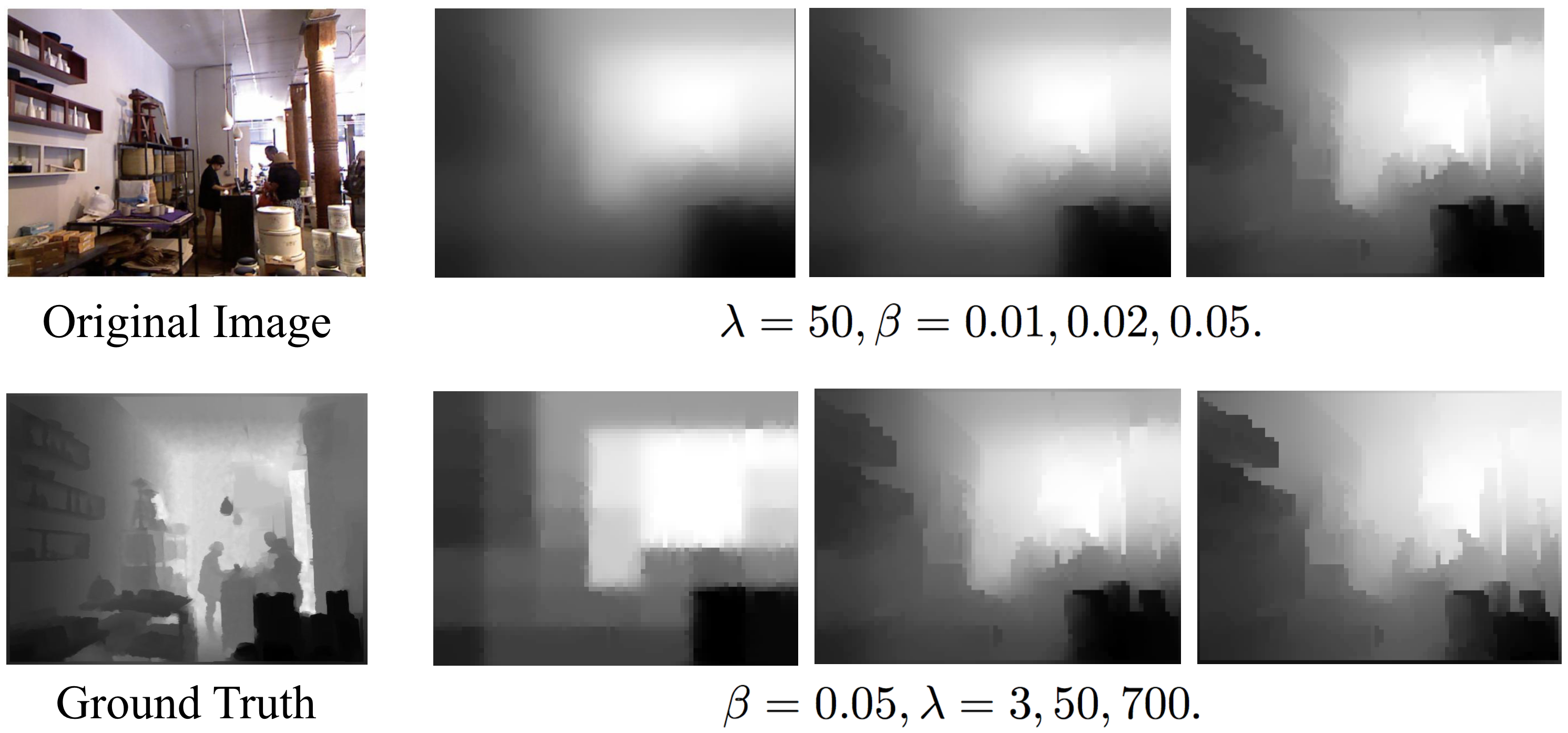}}
\end{center}
   \caption{Three depth maps in the upper row is generated when we fix $\lambda$ and increase $\beta$. Three depth maps in the bottom row is generated when we fix $\beta$ and increase $\lambda$.}
\label{fig:short}
\end{figure*} 

\subsection {Implementation Details}
Due to the high computational expense to optimize the CRF (for a $H \times W$ depth map, we have to calculate inverse of a $HW \times HW$ matrix), the final depth map cannot have high resolution. After experiment, we find that $H=63, W=84$ is a good tradeoff between resolution and running time. Thus after getting the input, we resize the rgb image to $63 \times 84$ and solve a CRF with $63 \times 84 = 5292$ variables. The image intensity is calculated from the rgb value of resized image.

\section{Experiment}
\subsection{Comparison with hand-crafted depth}
In this subsection, we design an experiment to compare depth calculated by our method with hand-crafted depth. 
\subsubsection{Evaluation procedure}
\text{\ \ \ \ }We use the following experiment procedure. For an arbitrary image, firstly we randomly pick $N$ (in our experiment, we take $N = 10$) points in image and ask human to directly label the depth of these points. Then we generate the depth map for the image using our method and write down the depth for the same $N$ points. We compare the depth of $N$ chosen points directly labeled by human and generated using our algorithm by following three metrics.

\subsubsection{Evaluation metrics}
\text{\ \ \ } Three metrics are used to evaluate the estimation result, \ie mean square error, cosine similarity with ground truth and pairwise rank accuracy. We use ground truth as our target. The formal definitions of these metrics are listed as follows:

\begin{itemize}
\item mean square error:
$\frac{1}{N} \sum_p{(d_p^{gt} - d_p)^2} $;
\item mean cosine similarity:
$ \frac{\bf{x}^{gt} \cdot \bf{x}}{|\bf{x}^{gt}| \cdot |\bf{x}|} $;
\item pairwise rank accuracy:\\
\begin{align*}
\frac{1}{N \cdot (N-1) / 2} &\sum_{p < p'}\mathbb{I} \{(d_p \leqslant d_{p'} \ and \ d_p^{gt} \leqslant d_{p'}^{gt}) \\
 &\ \ \ \ \ \  or\  (d_p \geqslant d_{p'} \ and \ d_p^{gt} \geqslant d_{p'}^{gt}) \};
\end{align*} 
\end{itemize}
where $N$ is the total number of pixels in all the evaluated image, $d_p^{gt}$ and $d_p$ are the ground truth and predicted depth respectively at pixel indexed by $p$, and $\mathbf{x}^{gt}$ and $\mathbf{x}$ are $T$-dimensional vectors whose components are $d_p^{gt}$ and $d_p$, respectively.

\subsection{Performance on NYU Depth dataset}
\begin{table}
\begin{center}
\begin{tabular}{|l|c|c|}
\hline
Metric             & Size to Depth & Hand-crafted Depth\\ 
\hline
Mean Square Error  & 0.016     & 0.048 \\
Cosines Similarity     & 0.976    & 0.935 \\
Pairwise Accuracy  & 0.858     & 0.818 \\
\hline
\end{tabular}
\end{center}
\caption{Result on three metrics. Depth generated by our method outperforms hand-crafted depth.}
\end{table}

We test images from NYU Depth dataset \cite{silberman11indoor, Silberman:ECCV12} and use our algorithm to generate depth map for them. Processing a standard image with size $480 \times 640$ takes tens of seconds. Some of the results are shown in Figure 2. Our labeling scheme performs consistently better than directly labeling depth in all the three metrics.
\subsection{Effect of hyperparameters}
In our algorithm, the strength of local connectivity is controlled by two hyper-parameters $\lambda \text{ and } \beta$.

$\lambda$ controls the continuity by trading off between CRF unary term and binary term.
\begin{align}
E(\mathbf{x}, \mathbf{y}) &= \sum_{p \in P}E_{unary}(\mathbf{y}_{p}, \mathbf{x}) \notag \\ &\quad + \lambda \sum_{(b, c) \in A} \frac{1}{2} E_{binary}(\mathbf{y}_b, \mathbf{y}_c, \mathbf{x})
\end{align} 

For smaller $\lambda$, the model cares more about matching depth with the annotated depth, despite slight discontinuity occurs. And for bigger $\lambda$, the model lays more importance on local continuity, even if it will cause mismatch with the annotated depth.

$\beta$ controls continuity by influencing weight of penalty term for depth gap.
\begin{equation}
\text{sim}(b, c) = e^{-\beta |I_b-I_c|}
\end{equation}
\begin{equation}
E_{binary}(\mathbf{y}_b, \mathbf{y}_c, \mathbf{x}) = \text{sim}(b, c) (\mathbf{y}_b-\mathbf{y}_c)^2
\end{equation}

$\text{sim}(b, c)$ is specially designed to have this form. The minus sign gives higher weight for smaller intensity difference. It basically says that the more similar two pixels are in rgb, the greater penalty they will get for same depth difference. The effect is, the more similar adjacent pixels are in rgb, the more similar they will tend to be in depth.

Figure 3 illustrates the effect of $\lambda$ and $\beta$.

\section{Conclusion}
We have presented a size to depth method for single monocular image depth estimation. The proposed method exploit the strength of human to estimate object size.  We explain the superiors of human labeling size compared to labeling depth. We show that the size labeled by human can be transferred to depth information through geometric relationships. A specific CRF model then propagate depth to generate the whole depth map. Experiments show that our method outperforms traditional depth-labeling method and generates quite satisfactory depth maps.
\section{Future Works}
\subsection{Faster alternative for manual labeling}
Although patch-to-size greatly reduces annotation effort, one still have to label like 7 by 7 patches which takes 3-5 minutes. Additional efforts can be done to further accelerate the labeling process. Assigning a single real-valued depth to image patches is an obvious application of CNNs. We are pretty sure that CNNs can be used to speedup this process. 
\subsection{Taking larger neighborhood into consideration}
In this paper, the weight of penalty term for depth gap only considers rgb of the two adjacent pixels. This might work in some case, but for adjacent pixels with same color but different depth or with different color but same depth, it will tend to give the right depth relatively high penalty. This is because that it only takes into account the two pixels. Other neighboring pixels are likely to also provide clue of whether their depth should be different or same. Ignoring such information might end up with erroneous solution. So the algorithm will hopefully work better by designing a function to derive penalty weight while considering larger neighborhood. A possible solution is to use a CNN as a projector from image patch space to lower dimension feature space. And train the CNN so that Euclidean distance of feature vector can act as a good similarity measure.

{\small
\bibliographystyle{ieee}
\bibliography{egbib}
}

\end{document}